\documentclass[10pt,twocolumn,letterpaper]{article}

\usepackage{cvpr}
\usepackage{times}
\usepackage{epsfig}
\usepackage{graphicx}
\usepackage{amsmath}
\usepackage{amssymb}

\usepackage[square, numbers, sort, comma]{natbib}
\usepackage{multirow}

\usepackage[breaklinks=true,bookmarks=false]{hyperref}

\cvprfinalcopy 

\def\cvprPaperID{****} 

\begin{document}

\title{Monocular Depth Estimation: A Survey}

\author{Amlaan Bhoi\\
Department of Computer Science\\
University of Illinois at Chicago\\
{\tt\small abhoi3@uic.edu}
}

\maketitle

\begin{abstract}
   Monocular depth estimation is often described as an ill-posed and inherently ambiguous problem. Estimating depth from 2D images is a crucial step in scene reconstruction, 3D object recognition, segmentation, and detection. The problem can be framed as: given a single RGB image as input, predict a dense depth map for each pixel. This problem is worsened by the fact that most scenes have large texture and structural variations, object occlusions, and rich geometric detailing. All these factors contribute to difficulty in accurate depth estimation. In this paper, we review five papers that attempt to solve the depth estimation problem with various techniques including supervised, weakly-supervised, and unsupervised learning techniques. We then compare these papers and understand the improvements made over one another. Finally, we explore potential improvements that can aid to better solve this problem.
\end{abstract}

\section{Introduction}

Estimating depth of a scene from a single image is an easy task for humans, but is notoriously difficult for computational models to do with high accuracy and low resource requirements. \textbf{Monocular Depth Estimation} (abbr. as \textit{MDE} hereafter) is this very task of estimating depth from a single RGB image. There are various advantages to be able to estimate depth from a single image. Some applications include scene understanding, 3D modelling, robotics, autonomous driving, \textit{etc}. Recovering depth information in these applications is more important when no other information such as stereo images, optical flow, or point clouds are unavailable.

Humans do well in this task becausee we can exploit features such as perspective, scale relative to known objects, appearance in lighting or occlusion, and more. MDE for computational model is an ill-posed problem because a single 2D image may be produced from an infinite number of distinct 3D scenes. To overcome this, earlier approaches resorted to exploiting those same statistically meaningful monocular cues or features such as perspective and texture information, object sizes, object localization, and occlusions.


In previous years, there has been significant work in depth estimation from stereo images or video sequences as seen in \cite{Ha2016, Kong2015, Karsch2015, Rajagopalan2004}. However, this requires more resources and data when compared to monocular depth estimation. In more recent years, MDE has gained traction and more deep learning models were developed for this problem \cite{Eigen2014, Eigen2015, Liu2015, Porzi2017}. All of these algorithms do not rely on hand-crafted features and are mostly deep convolutional neural networks. Based on their superior results, it is apparent learning deep features is superior to hand-crafted features.

\begin{figure}
  \centering
    \reflectbox{%
      \includegraphics[width=0.45\textwidth]{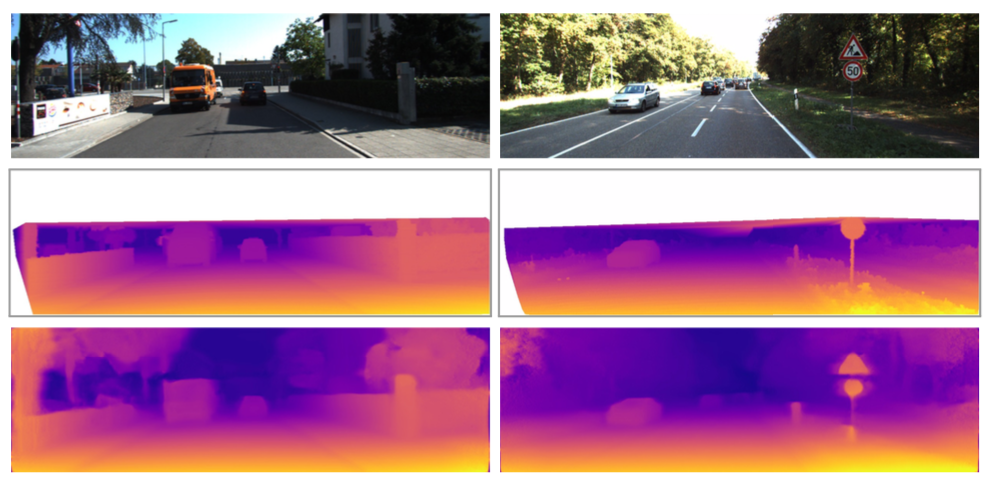}}
  \caption{Example of depth prediction by \citet{godard2017unsupervised} on KITTI \cite{Uhrig2017THREEDV} dataset.}
\end{figure}

More recent work also focus on combining information from multiple scales to better predict pixel-level depth \cite{Eigen2015, Xie2017, xu2018structured}. This can be done in multiple ways including \textit{fusing} feature maps corresponding to different layers in an architecture or aggregating the feature maps and using dimensionality reduction for regression. There has also been significant work using \textit{conditional random fields} (CRFs) \cite{xu2018structured, Chen2018, Li2015, Xu2018monocular} to greatly enhance predictions and yield further improvements.

\section{Problem Definition}

A standard problem formulation for monocular depth estimation can be stated as follows: Assuming the availability of a large training set of RGB-depth pair images, monocular depth estimation from single images can be regarded as a \textit{pixel-level continuous regression} problem. Due to the fact that this is a regression problem, a standard mean squared error (MSE) loss in log-space or its variants is usually employed as the loss function. This assumption is usually constrained to supervised learning where the task is pixel-wise continuous regression.

More specifically, the problem of predicting a depth map from a single RGB image can be viewed as following. Let $\mathcal{I}$ be the space of RGB images and $\mathcal{D}$ the domain of real-valued depth maps. Given a training set $\mathcal{T} = \{(\mathbf{I}_i, \mathbf{D}_i)\}^{M}_{i=1}, \mathbf{I}_i \in \mathcal{I}$ and $\mathbf{D}_i \in \mathcal{D}$, the task is to learn a non-linear mapping $\Phi:\mathcal{I}\rightarrow \mathcal{D}$. This formulation is applicable to supervised learning algorithms where pixel-level ground truth is available. Some methods relax this constraint by introducing different requirements and constraints. We shall mention these where applicable.

\section{Depth Map Prediction from a Single Image using a Multi-Scale Deep Network}

One of the earliest work inspiring the modern trends of depth estimation is by \citet{Eigen2014}. Modern approaches require the use of multi-scale features from a deep CNN \cite{Xie2017, xu2018structured, Eigen2015}. The introduction of this approach to use \textbf{multi-scale information} was introduced in this paper. This paper also introduced the concept of \textit{directly regressing} over pixels for depth estimation. Their network has two components: one that first estimates the global structure of the scene, then a second that refines it using local information. They use a special \textit{scale-invariant loss} to account for scale-dependent error.

\subsection{Model Architecture}

\textbf{Global Coarse-Scale Network. }The global coarse-scale network predicts the overall depth map structure using a global view of the scene. The network takes care of things like vanishing points, object locations, and room alignment. It contains five feature extraction layers of \textit{conv} and \textit{max-pooling} followed by two \textit{FC} layers. The final output is at $1/4$-resolution compared to input.

\textbf{Local Fine-Scale Network. }The task of the local fine-scale network is to perform local refinements. This takes care of details such as objects and wall edges. This network has three \textit{conv} layers with coarse features concatenated to the second block.

\subsection{Scale-Invariant Error}

The \textit{scale-invariant error} measures the relationship between points in the scene, irrespective of the absolute global scale. From a predicted depth map $y$ and ground truth $y^*$, each with $n$ pixels index by $i$, the $scale-invariant mean squared error$ (in log space) is:
\begin{equation}
    D(y, y^*) = \frac{1}{2n}\sum_{i=1}^{n}(\log y_i - \log y_{i}^{*} + \alpha(y, y^*))^2
\end{equation}
where $\alpha(y, y^*)=\frac{1}{n}\sum_i(\log y_{i}^{*} - \log y_i)$ is the value of $\alpha$ that minimizes the error for a given $(y, y^*)$. On the NYU Depth V2 \cite{Silberman2012} dataset, this method achieved a 7.165 RMSE score and thresholded accuracy of 0.967 where $\delta < 1.25^3$.

\section{Multi-Scale Continuous CRFs as Sequential Deep Networks for MDE}

\citet{Xu2018monocular} introduce a novel framework which exploits multi-scale estimations derived from CNN inner semantic layers by structurally fusing them within a unified CNN-CRF framework. The same way \citet{Li2015} fused features from two-stream CNN networks, the authors here also extract intermediate layers at different scales and fuse them. The addition in this approach is an introduction of two types of continuous CRF implementations: multi-scale CRFs and a cascade of multiple CRFs at each scale level $L$. The authors contributions are as follows:
\begin{itemize}
    \item A novel approach for predicting depth maps from RGB inputs which exploits multi-scale estimations derived from CNN inner layers by fusing them within a CRF framework.
    \item A method to implement mean field (MF) updates as sequential deep models.
\end{itemize}

\subsection{Fusing side outputs with continuous CRFs}

Let us explore the two methods to fuse intermediate features within a CRF framework.

\subsubsection{Multi-scale CRFs}

Given an $LN$-dimensional vector $\hat{s}$ obtained by concatenating side output score maps $\{s_1,...,s_L\}$ and an $LN$-dimensional vector $\mathbf{d}$, the CRF modeling the conditional distribution is defined as:
\begin{equation}
    P(\mathbf{d}|\mathbf{\hat{s}}) = \frac{1}{Z(\mathbf{\hat{s}})}\exp\{-E(\mathbf{d}, \mathbf{\hat{s}})\}
\end{equation}
where $Z(\mathbf{\hat{s}}) = \int_{d}\exp-E(\mathbf{d}, \mathbf{\hat{s}})$ is the partition function. The sum of quadratic unary terms is defined as:
\begin{equation}
    \phi(d_{i}^{l}, \mathbf{\hat{s}})=(d_{i}^{l} - s_{i}^{l})^2
\end{equation}
where $s_{i}^{l}$ is the regressed depth value at pixel $i$ and scale $l$. The pairwise potentials describing the relationship between pairs of hidden variables $d_{i}^{l}$ and $d_{j}^{k}$ is defined as:
\begin{equation}
    \psi(d_{i}^{l},d_{i}^{k}) = \sum_{m=1}^{M}\beta_{m}w_{m}(i, j, l, k, \mathbf{r})(d_{i}^{l}-d_{i}^{k})^2
\end{equation}
The derivation for the mean-field updates can be found in the original paper.

\subsubsection{Cascade CRFs}

The cascade model is based on $L$ CRF models, each one at a specific scale $L$ which are progressively stacked to use only the previous scale as an input to define features at the next level. The similarity between observation $o_{l}^{i}$ and hidden depth value $d{i}^{l}$ is:
\begin{equation}
    \phi(y_{i}^{l}, \mathbf{o}^{l}) = (d_{i}^{l}-o_{i}^{l})^2
\end{equation}
where $o_{i}^{l}$ is obtained combining regressed depth from previous outputs $\mathbf{s^l}$ and map $\mathbf{d}^{l-1}$ estimated by CRF at previous scale. The pairwise potentials used to enforce neighboring pixels with similar appearance to have close depth values area:
\begin{equation}
    \psi(d_{i}^{l}, d_{j}^{l})=\sum_{m=1}^{M}\beta_{m}K_{m}^{ij}(d_{i}^{l}-d{j}^{l})^2
\end{equation}
where $M=2$ are the Gaussian kernels; one for appearance features and other for pixel positions. More details to implement this CRF framework as a deep sequential model can be found in the original paper.

On NYU Depth V2 dataset \cite{Silberman2012}, with a ResNet50 \cite{he2016deep} as the front-end CNN, the authors achieved a 0.586 rms score and thresholded accuracy of 0.987 where $\delta < 1.25^3$.

\begin{table*}[t]
\centering
\begin{tabular}{|c|c|c|c|c|}
\hline
\textbf{Method}    & \textbf{Network Architecture} & \textbf{Multi-Scale Features} & \textbf{CRFs} & \textbf{Learning Paradigm} \\ \hline
\textbf{Multi-Scale CNN \citet{Eigen2014}}  & Deep CNN                      & Yes                           & No            & Supervised                 \\ \hline
\textbf{Multi-Scale CRFs \citet{Xu2018monocular}} & Deep CNN                      & Yes                           & Yes           & Supervised                 \\ \hline
\textbf{SAGCNF \citet{xu2018structured}}              & Deep CNN                      & Yes                           & Yes           & Supervised                 \\ \hline
\textbf{DORN \citet{fu2018deep}}             & DSE+SUM       & Yes                           & No            & Supervised                 \\ \hline
\textbf{LR \citet{godard2017unsupervised}}   & DispNet \cite{mayer2016large}                       & Yes; Only 4 scales                  & No            & Unsupervised               \\ \hline
\end{tabular}
\caption{Comparison of methods for Monocular Depth Estimation}
\end{table*}

\section{Structured Attention Guided Convolutional Neural Fields for MDE}

\citet{xu2018structured} introduce a framework extremely similar to the one by \citet{Xu2018monocular}. The authors' framework uses the same overall structure of having a front-end CNN architecture from which multi-scale information is extracted and fed into a \textit{continuous CRF} model. The major addition enforcement of similarity constraints and usage of \textbf{structured attention model} which can automatically regulate amount of information transferred between corresponding features at different scales. Their main contributions are:
\begin{itemize}
    \item Different from previous methods, this method does not consider as input only prediction maps but operates directly at feature-level. This method also claims to make faster inference than previous approaches.
    \item To robustly fuse features derived from multiple scales and enforce structure, the authors use a novel attention mechanism.
\end{itemize}
The primary idea of having an attention model is to control the flow of information. More specifically, an attention model $A=\{A_s\}_{s=1}^{S-1}$ parameterized by binary variables $\mathbf{A}_s=\{a_{s}^{i}\}_{i=1}^{N}, a_{s}^{i} \in \{0, 1\}$ is introduced. The attention variable $a_{s}^{i}$ regulates information which is allowed to flow between intermediate scale $s$ and final scale $S$ for pixel $i$.

\subsection{Structured Attention Guided Multi-Scale CRF}

Given observed multi-scale feature maps $\mathbf{X}$, we can estimate the latent multi-scale representations $\mathbf{Y}$ and attention variables $\mathbf{A}$ by modeling a CRF as:
\begin{equation}
    E(\mathbf{Y}, \mathbf{A}) = \Phi(\mathbf{Y}, \mathbf{A}) + \Xi(\mathbf{Y}, \mathbf{A}) + \Psi(\mathbf{A})
\end{equation}
where the first term is the sum of unary potentials, second term models the relationship between latent features at last scale with those of each intermediate scale, and the third term aims to enforce some structural constraints among attention variables. The sum of unary potentials is calculated as:
\begin{equation}
    \Phi(\mathbf{Y}, \mathbf{X}) = \sum_{s=1}^{S}\sum_{i}\phi(\mathbf{y}_{s}^{i}, \mathbf{x}_{s}^{i}) = -\sum_{s=1}^{S}\sum_{i}\frac{1}{2}\left \| \mathbf{y}_s^{i}-\mathbf{x}_{s}^{i} \right \|
\end{equation}
The second term is defined as:
\begin{equation}
    \Xi(\mathbf{Y}, \mathbf{A})=\sum_{s \neq S}\sum_{i, j}\xi (a_{s}^{i}, \mathbf{y}_{s}^{i}, \mathbf{y}_{S}^{i})
\end{equation}
Both these terms are similar to the ones in \cite{Xu2018monocular} except for the addition of attention variables $a_{s}^{i}$. The new term that enforces structure between attention variables is defined as:
\begin{equation}
    \Psi(\mathbf{A})=\sum_{s \neq S}\sum_{i, j}\psi(a_{s}^{i}, a_{s}^{j})=\sum_{s \neq S}\sum_{i, j}\beta_{i, j}^{s}a_{s}^{i}a_{s}^{j}
\end{equation}
where $\beta_{i, j}^s$ are coefficients to be learned.

\subsection{Network Structure, Implementation, and Optimization}

In practice, to update each attention map $\mathbf{a}_s$, the following can be followed: (i) perform message passing from two associated feature maps $\mathbf{\bar{y}}_s$ and $\mathbf{\bar{y}}_S$ via convolutional operations $\mathbf{\hat{a}}_s \leftarrow \mathbf{y}_{s} \bigodot (\mathbf{K}_s \otimes \mathbf{\bar{y}}_S)$; (ii) perform message passing on attention map with $\mathbf{\tilde{a}} \leftarrow \beta_s \otimes \mathbf{\bar{a}}_s$; (iii) normalize with sigmoid functions $\mathbf{\bar{a}_s} \leftarrow \sigma(-(\mathbf{\hat{a}} \oplus \mathbf{\tilde{a}}_s))$. A similar procedure can be followed for mean-field updates for $\mathbf{y}_S$. To reduce overhead, the authors do not perform mean-field updates for intermediate scales.

The front-end CNN used is a ResNet-50 \cite{he2016deep} where the CRF is used to refine last scale feature map from semantic layer \textit{res5c}. Each convolutional block has the same number of channels. The kernel size for $\mathbf{K}_s$ and $\mathbf{\beta}_s$ is set to 3 with stride 1 and padding 1 to speed up calculation. For optimization, the square loss function is used:
\begin{equation}
    \begin{split}
    \mathcal{L}_F((\mathcal{I}, \mathcal{D}; \mathbf{\Theta}_{e}, \mathbf{\Theta}_{c}, \mathbf{\Theta}_{d}) = \\
    \sum_{i=1}^{M} \left \| F(\mathbf{I}_{i}^{l}; \mathbf{\Theta}_{e}, \mathbf{\Theta}_{c}, \mathbf{\Theta}_{d}) - \mathbf{D}_{i}^{l} \right \|)_{2}^{2}
    \end{split}
\end{equation}
The whole network is jointly optimized via backpropagation with standard stochastic gradient descent. On the NYU Depth V2 \cite{Silberman2012} dataset, this method achieved a 0.593 rms score and thresholded accuracy of 0.986 where $\delta < 1.25^3$.

\section{Deep Ordinal Regression Network for MDE}

Deviating from previous approaches to MDE, \citet{fu2018deep} approach monocular depth estimation by phrasing the problem as an ordinal regression task instead of continuous depth map prediction. The main contributions are:
\begin{itemize}
    \item Discretized continuous depth map into a number of intervals and cast depth network learning as an ordinal regression problem.
    \item Introduced a \textbf{space-increasing discretization} (SID) strategy instead of \textbf{uniform discretization} (UD) motivated by the fact that uncertainty in depth prediction increases along with the underlying ground-truth depth indicating that it would be better to allow a relatively large error when predicting a larger depth value to avoid over-strengthened influence of large depth values on training process.
\end{itemize}

\subsection{Space-Increasing Discretization}

Before we understand the network architecture and training details, let us peruse over how the space-increasing discretization works. To quantize a depth interval $[\alpha, \beta]$ into a set of representative discrete values, we normally use uniform discretization. The disadvantage for that strategy is it would induce an over-strengthened loss for large depth values. Based on that intuition, the authors propose the SID strategy. Assuming that a depth interval $[\alpha, \beta]$ needs to be discretized into $K$ sub-intervals, UD and SID can be formulated as:
\begin{equation}
    \begin{split}
    \textup{UD}: t_i = \alpha + (\beta - \alpha) * i/K \\
    \textup{SID}: t_i = e^{log(\alpha) + \frac{log(\beta/\alpha)*i}{K}}
    \end{split}
\end{equation}
where $t_i \in \{t_0, t_1,...,t_K\}$ are discretization thresholds.

\subsection{Network Architecture, Learning, and Inference}

\subsubsection{Network Architecture}

The network architecture is divided into two portions: a dense feature extractor and a scene understanding modular. The \textbf{dense feature extractor} in previous approaches have utilized a combination of max-pooling and striding that significantly reduces the spatial resolution of the feature maps. Some remedies include skip connections, stage-wise refinement, or multi-layer deconvolution network. However, these are just partial solutions and are not desirable. Thus, the authors remove the last few downsampling oeprators and insert subsequent \textit{conv} layers (dilated convolutions) to enlarge the field-of-view of filters without decreasing spatial resolution or increaasing number of parameters. The \textbf{scene understanding modular} consists of three parallel components: an \textit{atrous spatial pyramid pooling} (ASPP) module, a cross-channel learner, and a full-image encoder. The \textit{ASPP} module extracts features from multiple large receptive fields via dilated convolutional operations. The pure 1x1 convolutional branches (pointwise convolution) can learn complex cross-channel interactions. The full-image encoder captures global contextual information and can significantly clarify local confusions in depth estimation.

\subsubsection{Learning and Inference}

After discretizing depth values, it seems natural to turn the objective into a multi-class classification problem and adopt \textit{softmax} regression. However, this approach loses the importance of ordering. There is a strong ordinal correlation between discrete labels and form a well-defined set. Thus, the problem is casted as an ordinal regression problem. Let $\chi=\varphi(I, \Phi)$ denote the feature maps of size $W \times H \times C$ given an image $I$, where $\Phi$ are the parameters involved in the dense feature extractor and scene understanding modular. $Y = \psi(\chi, \Theta)$ of size $W \times H \times 2K$ denotes the ordinal outputs for each spatial locations, where $\Theta = \{\theta_0, \theta_1,...,\theta_{2K-1}\}$ contains weight vectors. And $l_{(w, h)} \in \{0,1,...,K-1\}$ is the discrete label produced by SID at spatial location $(w, h)$. The ordinal loss is defined as:
\begin{equation}
\begin{split}
    \mathcal(L)(\chi, \Theta) = -\frac{1}{\mathcal{N}}\sum_{w=0}^{W-1}\sum_{h=0}^{H-1}\Psi(w, h, \chi, \Theta) \\
\Psi(h, w, \chi, \Theta) = \sum_{k=0}^{l_{(w, h)-1}}\log(\mathcal{P}_{(w, h)}^{k}) + \\
\sum_{k=l_{(w, h)}}^{K-1}(\log(1-\mathcal{P}_{(w, h)}^{k})), \\
\mathcal{P}_{(w, h)}^{k} = P(\hat{l}_{(w, h)} > k|\chi, \Theta)
\end{split}
\end{equation}
where $\mathcal{N} = W \times H$ and $\hat{l}_{(w, h)}$ is the estimated discrete value decoding from $y_{(w, h)}$. The minimization of the loss function is done through an iterative optimization algorithm which can be found in the original paper. The authors present their results on KITTI \cite{Uhrig2017THREEDV}, NYU Depth V2 \cite{Silberman2012}, and Make3D \cite{saxena2009make3d} datasets. On the NYU Depth V2 dataset, they achieved a 0.509 rms score and thresholded accuracy of 0.992 where $\delta < 1.25^3$.

\begin{table*}[t]
\centering
\begin{tabular}{|c|c|c|c|}
\hline
\textbf{Dataset} & \textbf{Statistics} & \textbf{Annotation Available}        & \textbf{Scene} \\ \hline
NYUD-V2 \cite{Silberman2012}          & 1449 + 407K RAW     & Depth + Segmentation                 & Indoor         \\ \hline
KITTI \cite{Uhrig2017THREEDV}           & 94K Frames          & Depth aligned with RAW data          & Street         \\ \hline
Make3D \cite{saxena2009make3d}          & 500 Frames          & Depth                                & Outdoor        \\ \hline
\end{tabular}
\caption{Datasets for Monocular Depth Estimation}
\end{table*}

\section{Unsupervised MDE with Left-Right Consistency}

We take a final leap towards an unsupervised learning approach to MDE by \citet{godard2017unsupervised}. The authors employ binocular stereo footage with left and right images of a view in training but only use one view for testing and inference. By exploiting \textit{epipolar geometry constraints}, they generate disparity images by training their network with an image reconstruction loss. The model does not require any labelled depth data and learns to predict pixel-level correspondence between pairs of rectified stereo images. Their main contributions are:
\begin{itemize}
    \item A new network architecture that performs end-to-end unsupervised monocular depth estimation with a novel training loss to enforce left-right consistency inside the network.
    \item Results that show that the network generalizes well enough to at least three datasets they tested on.
\end{itemize}
The authors introduce a fully convolutional neural network that is loosely inspired by the supervised DispNet by \citet{mayer2016large}. However, just incorporating image reconstruction loss is not enough as it can result in good image reconstruction but poor depth quality. To remedy this, the authors introduce new losses to help train the network to predict depth with high accuracy.

\subsection{DE as Image Reconstruction, DE Network, and Training Loss}

Let us now see how to pose the depth estimation problem as an image reconstruction problem, explore the depth estimation network, and explore the novel training loss.

\subsubsection{Depth Estimation as Image Reconstruction}

Given a single image $I$, the goal is to learn a function $f$ that can perdict per-pixel scene depth $\hat{d} = f(I)$. Learning this function through supervised methods is not practical due to lack of precise ground truth labels. Expensive hardware such as laser scanners may also suffer from errors in scenes with movements and reflections. Thus, the authors propose to learn a function able to reconstruct one image from another. Specifically, at training time, the network is given two images $I^l$ and $I^r$ corresponding to the left and right color images from a calibrated stereo pair. The network attempts to find dense correspondence $d^r$ that when applied to the left image, can enable us to reconstruct the right image. The reconstructed image is denoted as $I^l(d^r)$ or $\tilde{I}^r$. Similarly, $\tilde{I}^l = I^r(d^l)$. Assuming images are rectified, $d$ corresponds to image disparity - a scalar value per pixel that the model will learn to predict. Given the baseline distance $b$ between the cameras and focal length $f$, we can trivially recover depth $\hat{d}$ from predicted disparity as $\hat{d}=bf/d$.

\subsubsection{Depth Estimation Network}

The network proposed by the authors generates the predicted image with backward mapping using a \textit{bilinear sampler}. Instead of sampling from left or right image individually and trying to predict disparity maps for one, \citet{godard2017unsupervised} train the model to predict both disparity maps using a novel left-right consistency loss. The fully convolutional network consists of an encoder and a decoder with skip connections. Disparity predictions are predicted at four scales. Even though one input is required in testing, the network still predicts two disparity maps.

\subsubsection{Loss Functions}

\textbf{Training Loss.} The authors define a loss $C_s$ at each output scale $s$, forming the loss as:
\begin{equation}
    C_s = \alpha_{ap}(C_{ap}^{l}+C_{ap}^{r} + \alpha_{ds}(C_{ds}^l+C_{ds}^r) + \alpha_{lr}(C_{lr}^l+C_{lr}^r))
\end{equation}
where $C_{ap}$ encourages similarity in reconstructed image, $C_{ds}$ enforces smoothness disparities, and $C_{lr}$ perfers predicted left and right disparities to be consistent.

\textbf{Appearance Match Loss.} The network learns to generate an image by sampling pixels from the opposite stereo image. In this paper, the image formation model uses the image sampler from the spatial transformer network (STN) \cite{jaderberg2015spatial} to sample input image using a disparity map. The $C_{ap}$ loss is a combination of an $L1$ and single scale SSIM  \cite{wang2004image} loss. This loss is defined as:
\begin{equation}
    C_{ap}^l = \frac{1}{N}\alpha\frac{1-\textup{SSIM}(I_{ij}^l, \tilde{I}_{ij}^l)}{2}+(1-\alpha)\left \| I_{ij}^l - \tilde{I}_{ij}^l \right \|
\end{equation}
Here, the authors use a simplified SSIM with a $3 \times 3$ block filter instead of a Gaussian, and set $\alpha=0.85$.

\textbf{Disparity Smoothness Loss.} Disparities are encouraged to be locally smooth with an $L1$ penalty on disparity gradients $\partial d$. Because depth may have discontinuity, this loss is weight with an edge-aware term using image gradients $\partial I$,
\begin{equation}
    C_{ds}^l = \frac{1}{N}|\partial_x d_{ij}^l|e - \left \| \partial_x I_{ij}^l \right \| + |\partial_y d_{ij}^l|e - \left \| \partial_y I_{ij}^l \right \|
\end{equation}

\textbf{Left-Right Disparity Consistency Loss.} To ensure coherence between the left and right disparity maps, the authors introduce an $L1$ left-right disparity consistency penalty. This cost attempts to make the left-view disparity map equal to the \textit{projected} right-view disparity map,
\begin{equation}
    C_{lr}^l = \frac{1}{N}\sum_{i, j}|d_{ij}^l - d_{ij + d_{ij}^l}^r|
\end{equation}
At test time, the network predicts disparity at the finest scale level for left image $d^l$. Using the camera distance and focal length, the disparity map can now be converted to a depth map. On the KITTI \cite{Uhrig2017THREEDV} dataset, this method achieved a 4.935 RMSE score and thresholded accuracy of 0.976 where $\delta < 1.25^3$.

\section{Datasets}

There are conventional and more modern and unconventional datasets available for monocular depth estimation today. We shall explore the conventional datasets here that can be used for evaluation and comparison against other works.

\textbf{NYUD-V2. }The NYU Depth V2 dataset was introduced at ECCV 2012 by \citet{Silberman2012}. There are \textbf{1449} densely labeled pairs of aligned RGB and depth images, \textbf{464} scenes taken from 3 cities, and \textbf{407K} unlabeled frames. It is common to down-sample images to speed up training. For example, \citet{fu2018deep} reduced the resolution of images to $288 \times 384$.

\textbf{KITTI. } The KITTI dataset by \cite{Uhrig2017THREEDV} was introduced in IJRR in 2013. Instead of utilizing the entire KITTI dataset, it is common to follow the \textit{Eigen split}. This split contains 23,488 images from 32 scenes for training and 697 images from 29 scenes for testing.

\textbf{Make3D. } The Make3D dataset \cite{saxena2009make3d} contains 534 outdoor images, 400 for training and 134 for testing. The resolution of these imagees are $2272 \times 1704$ while the ground truth depth map resolutions are $55 \times 305$. It is also common to down-sample these images to decrease training time.

\begin{table*}[ht]
\centering
\begin{tabular}{|c|c|c|c|c|c|c|}
\hline
\multirow{2}{*}{\textbf{Method}} & \multicolumn{3}{c|}{\textbf{Error (lower is better)}} & \multicolumn{3}{c|}{\textbf{Accuracy (higher is better)}}                                                                           \\ \cline{2-7} 
                                 & \textbf{rel}    & \textbf{log10}    & \textbf{rms}    & $\mathbf{\delta < 1.25}$ & $\mathbf{\delta < 1.25^2}$ & $\mathbf{\delta < 1.25^3}$ \\ \hline
\textbf{\citet{Eigen2014}}                   & 0.215           & -                 & 0.907               & 0.611                         & 0.887                                            & 0.971                                            \\ \hline
\textbf{\citet{Xu2018monocular}}                      & 0.121           & 0.052             & 0.586           & 0.811                         & 0.954                                            & 0.987                                            \\ \hline
\textbf{\citet{xu2018structured}}                     & 0.125           & 0.057             & 0.593           & 0.806                         & 0.952                                            & 0.986                                            \\ \hline
\textbf{\citet{fu2018deep}}                      & 0.115           & 0.051             & 0.509           & 0.828                         & 0.965                                            & 0.992                                            \\ \hline
\textbf{\citet{godard2017unsupervised}}                  & -               & -                 & -               & -                             & -                                                & -                                                \\ \hline
\end{tabular}
\caption{Comparison of methods' performance on \textbf{NYUD V2} \cite{Silberman2012} dataset.}
\end{table*}

\begin{table*}[ht]
\centering
\begin{tabular}{|c|c|c|c|c|c|c|}
\hline
\multirow{2}{*}{\textbf{Method}} & \multicolumn{3}{c|}{\textbf{Error (lower is better)}} & \multicolumn{3}{c|}{\textbf{Accuracy (higher is better)}}                                                                           \\ \cline{2-7} 
                                 & \textbf{rel}    & \textbf{sq rel}    & \textbf{rms}    & $\mathbf{\delta < 1.25}$ & $\mathbf{\delta < 1.25^2}$ & $\mathbf{\delta < 1.25^3}$ \\ \hline
\textbf{\citet{Eigen2014}}                   & 0.190           & 1.515                 & 7.156               & 0.692                         & 0.899                                            & 0.967                                            \\ \hline
\textbf{\citet{Xu2018monocular}}                      & -           & -             & -           & -                         & -                                            & -                                            \\ \hline
\textbf{\citet{xu2018structured}}                     & 0.122           & 0.897             & 4.677           & 0.818                         & 0.954                                            & 0.985                                            \\ \hline
\textbf{\citet{fu2018deep}}                      & 0.071           & 0.268             & 2.271           & 0.936                         & 0.985                                            & 0.995                                            \\ \hline
\textbf{\citet{godard2017unsupervised}}                  & 0.108               & 0.657                 & 3.729               & 0.873                             & 0.954                                                & 0.979                                                \\ \hline
\end{tabular}
\caption{Comparison of methods' performance on \textbf{KITTI} \cite{Uhrig2017THREEDV} dataset.}
\end{table*}

\section{Discussion}

Let us now explore how the approaches taken over different papers evolved and what are the similarities and differences in certain aspects of their approaches. In such progression, we get an idea about possible future scope of work in this area.

We first start with \textbf{Depth Map Prediction from a Single Imageusing a Multi-Scale Deep Network} by \citet{Eigen2014} who first proposed \textit{directly regressing} over pixel values and using \textit{multi-scale features} for monocular depth estimation. Their use of \textit{global-coarse scale network} and \textit{local fine-scale network} is the foundation upon which more modern techniques utilize multi-scale features to accomplish the same tasks of global depth prediction and local refinements. They also introduce \textit{scale-invariant loss} which helps mitigate errors in depth due to scale. Similar loss methods are utilized in future work as we shall see.

We now move on to \textbf{Multi-Scale Continuous CRFs as Sequential Deep Networks for Monocular Depth Estimation} by \citet{Xu2018monocular} who first propose using \textit{continuous CRFs} as deep neural networks for depth regression over pixels. They extend the idea of using \textit{multi-scale features} and introduce \textit{conditional random fields} as sequential deep network implementations. Instead of running a CRF as a shallow model over image features, they implement multi-scale CRFs, cascading CRFs, and \textit{mean-field updates} as sequential deep networks. The goal is to basically fuse the multi-scale features using CRFs to learn the structure of data over multiple scales.

The third approach we tackle builds upon the previous approach by adding \textit{attention variables}. \textbf{Structured Attention Guided Convolutional Neural Fields for Monocular Depth Estimation} by \citet{xu2018structured} extends the idea of fusing multi-scale features using CRFs by integrating attention mechanism. As with the previous method, they implement their CRF framework within an \textit{encoder-decoder} architecture to enable end-to-end training of the network. The addition of attention variables ensure the CRFs control the flow of information between different scales to avoid unnecessary updates.

The fourth approach called \textbf{Deep Ordinal Regression Network for Monocular Depth Estimation} by \citet{fu2018deep} takes a different approach to monocular depth estimation. Instead of regressing over the depth value per pixel, the authors introduced an \textit{ordinal regression} scheme using \textit{spacing-increasing strategy}. This helps quantize the outputs to intervals. They have a \textit{dense feature extractor} (front-end CNN) and \textit{scene understanding modular}. This approach also foregoes the encoder-decoder scheme, does not use MSE as error metric, and avoids pooling to avoid reducing resolution. The new addition is using \textit{dilated convolutions} and full-image encoders.

The final paper we explored is \textbf{Unsupervised Monocular Depth Estimation with Left-Right Consistency} by \cite{godard2017unsupervised}. The authors completely change the problem scope and introduce new inputs. They use two images (left \& right) in training. The objective is, in testing, given one image view, reconstruct the other image's disparity map and construct the corresponding depth map. They use \textit{appearance matching loss}, \textit{disparity smoothness loss}, and \textit{left-right disparity consistency loss}. This method is completely unsupervised as it does not require depth map as input during training or inference.

For supervised learning methods, multi-scale feature extraction is definitely needed for decent performance. More improvements can be made to the overall structure and framework. For unsupervised learning methods, more features or aspects of the problem is required. For example, \citet{mahjourian2018unsupervised} use \textit{camera egomotion} and \textit{3D geometric constraints} (iterative closest point).

In conclusion, we explored five methods for monocular depth estimation with various improvements and pitfalls. There are many more approaches to this problem that we can explore. This problem can still be solved more efficiently and with more precision.


{\small
\bibliographystyle{ieee}
\bibliography{egbib}
}

\end{document}